\documentclass[letterpaper,10pt,conference]{ieeeconf}

\IEEEoverridecommandlockouts                            
\usepackage{color,xcolor}
\usepackage{epsfig}
\usepackage{graphicx}

\usepackage{adjustbox}
\usepackage{array}
\usepackage{booktabs}
\usepackage{colortbl}
\usepackage{float,wrapfig}
\usepackage{hhline}
\usepackage{multirow}
\usepackage{subcaption} % issues a warning with CVPR/ICCV format
\usepackage{amsmath,amsfonts}
\usepackage{bm}
\usepackage{nicefrac}
\usepackage{microtype}
\usepackage{inconsolata}
\usepackage{pifont}

\usepackage{changepage}
\usepackage{extramarks}
\usepackage{fancyhdr}
\usepackage{lastpage}
\usepackage{setspace}
\usepackage{soul}
\usepackage{xspace}
\usepackage{indentfirst}
\usepackage{paralist}
\usepackage{dcolumn,booktabs}
\newcolumntype{d}[1]{D{.}{.}{#1}}

\usepackage[pagebackref=true,breaklinks=true,letterpaper=true,colorlinks,citecolor=gray,bookmarks=false]{hyperref}
\usepackage{url}

\usepackage{algorithm, algorithmic}
\usepackage{enumerate}
\usepackage{lipsum}
\newcolumntype{L}[1]{>{\raggedright\let\newline\\\arraybackslash\hspace{0pt}}m{#1}}
\newcolumntype{C}[1]{>{\centering\let\newline\\\arraybackslash\hspace{0pt}}m{#1}}
\newcolumntype{R}[1]{>{\raggedleft\let\newline\\\arraybackslash\hspace{0pt}}m{#1}}

\newcommand{\sect}[1]{Section~\ref{#1}}

\newcommand{\fig}[1]{Figure~\ref{#1}}

\newcommand{\tab}[1]{Table~\ref{#1}}

\newcommand{\ignorethis}[1]{}

\def\naive{na\"{\i}ve\xspace}

\makeatletter
\DeclareRobustCommand\onedot{\futurelet\@let@token\@onedot}
\def\@onedot{\ifx\@let@token.\else.\null\fi\xspace}

\def\eg{\emph{e.g}\onedot} 
\def\ie{\emph{i.e}\onedot}

\makeatother

\definecolor{citecolor}{RGB}{34,139,34}
\definecolor{mydarkblue}{rgb}{0,0.08,1}
\definecolor{mydarkgreen}{rgb}{0.02,0.6,0.02}
\definecolor{mydarkred}{rgb}{0.8,0.02,0.02}
\definecolor{mydarkorange}{rgb}{0.40,0.2,0.02}
\definecolor{mypurple}{RGB}{111,0,255}
\definecolor{myred}{rgb}{1.0,0.0,0.0}
\definecolor{mygold}{rgb}{0.75,0.6,0.12}
\definecolor{myblue}{rgb}{0,0.2,0.8}
\definecolor{mydarkgray}{rgb}{0.66,0.66,0.66}

\newcommand{\myparagraph}[1]{\vspace{2.5pt}\noindent\textbf{#1}~}

\def\model{BEVFusion\xspace}
\definecolor{linkcolor}{RGB}{255,0,0}
\definecolor{revisioncolor}{RGB}{0,0,0}

\title{\LARGE \bf
BEVFusion: Multi-Task Multi-Sensor Fusion \\ with Unified Bird's-Eye View Representation
}

\begin{document}

\author{Zhijian Liu$^{*,1}$, Haotian Tang$^{*,1}$, Alexander Amini$^{1}$, Xinyu Yang$^{1}$, Huizi Mao$^{2}$, Daniela L. Rus$^{1}$, Song Han$^{1}$
\thanks{$*$ The first two authors contributed equally and are listed alphabetically. This work was supported by MIT-IBM Watson AI Lab, National Science Foundation, Hyundai Motor, Qualcomm, NVIDIA and Apple. Zhijian Liu was partially supported by the Qualcomm Innovation Fellowship.}%
\thanks{$^{1}$ Z. Liu, H. Tang, A. Amini, X. Yang, D. Rus, and S. Han are with Massachusetts Institute of Technology, Cambridge, MA 02139, USA.}%
\thanks{$^{2}$ H. Mao is with OmniML, San Jose, CA 95131, USA.}%
}

\maketitle

\begin{abstract}

Multi-sensor fusion is essential for an accurate and reliable autonomous driving system. Recent approaches are based on point-level fusion: augmenting the LiDAR point cloud with camera features. However, the camera-to-LiDAR projection throws away the semantic density of camera features, hindering the effectiveness of such methods, especially for semantic-oriented tasks (such as 3D scene segmentation). In this paper, we propose BEVFusion, an efficient and generic multi-task multi-sensor fusion framework. It unifies multi-modal features in the shared bird's-eye view (BEV) representation space, which nicely preserves both geometric and semantic information. To achieve this, we diagnose and lift the key efficiency bottlenecks in the view transformation with optimized BEV pooling, reducing latency by more than \textbf{40$\times$}. BEVFusion is fundamentally task-agnostic and seamlessly supports different 3D perception tasks with almost no architectural changes. It establishes the new state of the art on the nuScenes benchmark, achieving \textbf{1.3\%} higher mAP and NDS on 3D object detection and \textbf{13.6\%} higher mIoU on BEV map segmentation, with \textbf{1.9$\times$} lower computation cost. Code to reproduce our results is available at \url{https://github.com/mit-han-lab/bevfusion}.

\end{abstract}
\section{Introduction}
\label{sect:intro}

Autonomous driving systems are equipped with diverse sensors. For instance, Waymo's self-driving vehicles have 29 cameras, 6 radars, and 5 LiDARs. Different sensors provide complementary signals: \eg, cameras capture rich semantic information, LiDARs provide accurate spatial information, while radars offer instant velocity estimation. Thus, multi-sensor fusion is essential for accurate and reliable perception.

\begin{figure}[t]
\centering
\includegraphics[width=\linewidth]{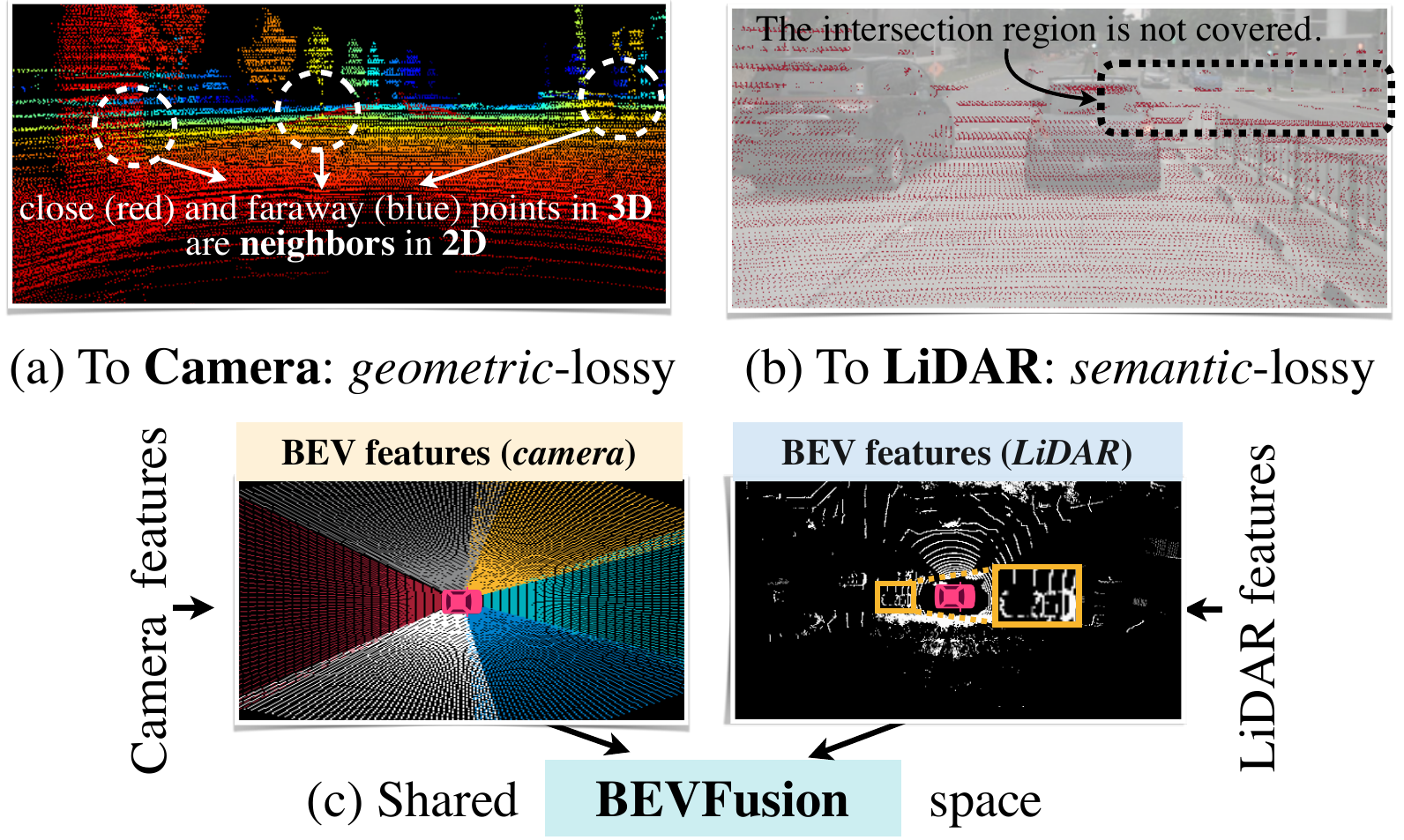}
\caption{\model unifies camera and LiDAR features in a \textit{shared} BEV space instead of mapping one modality to the other. It preserves camera's \textit{semantic density} and LiDAR's \textit{geometric structure}.}% \SH{for (c), we can add a car in the middle of the camera BEV figure, facing right, and remove ``front'' and ``back''}}
\label{fig:teaser}
\vspace{-20pt}
\end{figure}

Data from different sensors are expressed in fundamentally different modalities: \eg, cameras capture data in perspective view and LiDAR in 3D view. To resolve this view discrepancy, we have to find a \emph{unified representation} that is suitable for multi-task multi-modal feature fusion. Due to the tremendous success in 2D perception, the natural idea is to project the LiDAR point cloud onto the camera and process the RGB-D data with 2D CNNs. However, this LiDAR-to-camera projection introduces severe geometric distortion (see \fig{fig:teaser}\textcolor{linkcolor}{a}), which makes it less effective for geometric-oriented tasks, such as 3D object recognition.

Recent sensor fusion methods follow the other direction. They augment the LiDAR point cloud with semantic labels~\cite{vora2020pointpainting}, CNN features~\cite{wang2021pointaugmenting,li2022deepfusion} or virtual points from 2D images~\cite{yin2021multimodal}, and then apply an existing LiDAR-based detector to predict 3D bounding boxes. Although they have demonstrated remarkable performance on large-scale detection benchmarks, these point-level fusion methods barely work on semantic-oriented tasks, such as BEV map segmentation~\cite{pan2019crossview,philion2020lift,li2021hdmapnet,zhou2022cross}. This is because the camera-to-LiDAR projection is semantically lossy (see \fig{fig:teaser}\textcolor{linkcolor}{b}): for a typical 32-beam LiDAR scanner, only 5\% camera features will be matched to a LiDAR point while all others will be dropped. Such density differences will become even more drastic for sparser LiDARs (or radars).

In this paper, we propose BEVFusion to unify multi-modal features in a shared bird's-eye view (BEV) representation space for task-agnostic learning. We maintain both geometric structure and semantic density (see \fig{fig:teaser}\textcolor{linkcolor}{c}) and naturally support most 3D perception tasks (since their output space can be naturally captured in BEV). While converting all features to BEV, we identify the major prohibitive efficiency bottleneck in the view transformation: \ie, the BEV pooling operation alone takes more than 80\% of the model's runtime. Then, we propose a specialized kernel with precomputation and interval reduction to eliminate this bottleneck, achieving more than \textbf{40$\times$} speedup. Finally, we apply the fully-convolutional BEV encoder to fuse the unified BEV features and append a few task-specific heads to support different target tasks.
\begin{figure*}[ht]
\centering
\includegraphics[width=\linewidth]{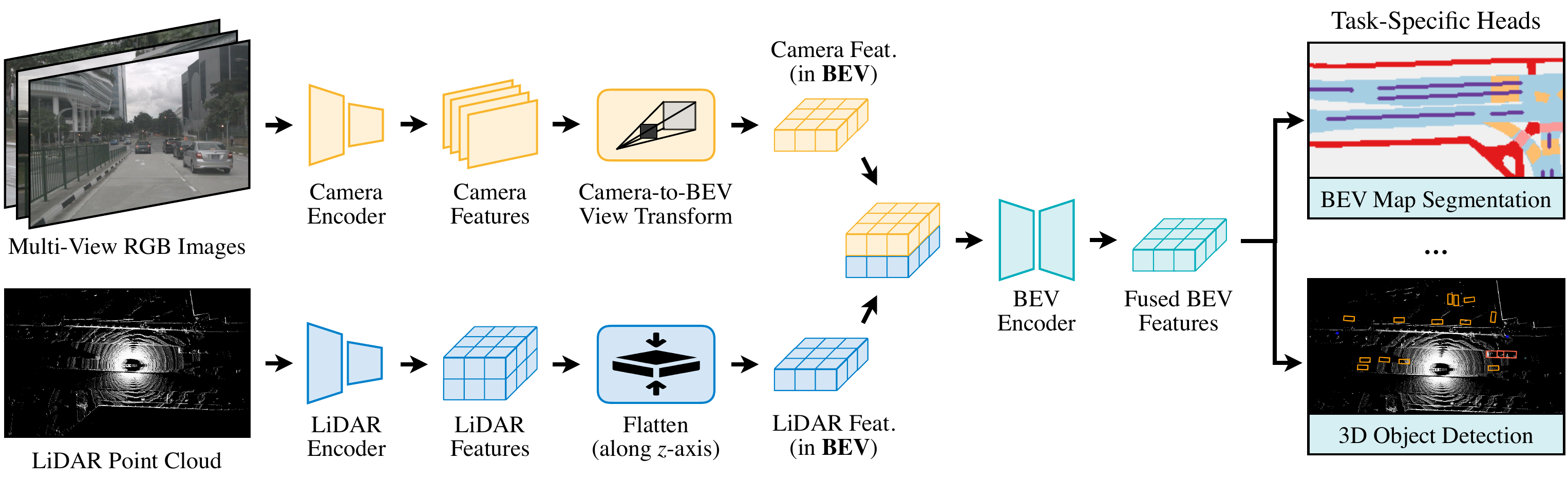}
\caption{\textbf{BEVFusion} extracts features from multi-modal inputs and converts them into a shared bird's-eye view (BEV) space efficiently using view transformations. It fuses the unified BEV features with a fully-convolutional BEV encoder and supports different tasks with task-specific heads.}
\label{fig:method:overview}
\vspace{-15pt}
\end{figure*}

BEVFusion sets the new state-of-the-art 3D object detection performance on both nuScenes and Waymo benchmarks. It outperforms all published methods with or without test-time augmentation and model ensemble.
% It outperforms point-level sensor fusion methods that are tailored for this type of geometric-oriented task.
BEVFusion demonstrates even more significant improvements on BEV map segmentation. It achieves \textbf{6\%} higher mIoU than camera-only models and \textbf{13.6\%} higher mIoU than LiDAR-only models, while existing fusion methods hardly work. Moreover, BEVFusion is highly efficient, delivering all these results with \textbf{1.9$\times$} lower computation cost.

While point-level fusion has been the go-to choice over the past three years, BEVFusion provides a fresh perspective to the field of multi-sensor fusion by rethinking ``\emph{Is LiDAR space the right place to perform sensor fusion?}''. It showcases the superior performance of an alternative paradigm that has been previously overlooked.
% We believe that our pilot study will arouse more attention and exploration into this important question.
% BEVFusion breaks the long-standing belief that point-level fusion is the best solution to multi-sensor fusion.
Simplicity is also its key strength. We hope this work will serve as a simple yet strong baseline for future sensor fusion research and inspire the researchers to rethink the design and paradigm for generic multi-task multi-sensor fusion. %We will make the code available to the public.
\section{Related Work}

\myparagraph{LiDAR-Based 3D Perception.}
Researchers have designed single-stage 3D object detectors~\cite{zhou2018voxelnet,lang2019pointpillars,yan2018second,zhu2019classbalanced,yang20203dssd,zhou2019mvf} that extract flattened point cloud features using PointNets~\cite{qi2017pointnet++} or SparseConvNet~\cite{graham20183d} and perform detection in the BEV space. Later, \cite{yin2021center,ge2021afdet,chen2020object,qi2021offboard,fan2021rangedet,chen2021polarstream,wang2021object} explore anchor-free single-stage 3D object detection. Another stream of research \cite{shi2019pointrcnn,chen2019fast,shi2019part,shi2019pvrcnn,shi2021pvrcnn++,li2021lidar} focuses on two-stage object detector design, which adds an RCNN network to existing one-stage object detectors.

\myparagraph{Camera-Based 3D Perception.}
Due to the high cost of LiDAR sensors, researchers spend significant efforts on camera-only 3D perception. FCOS3D~\cite{wang2021fcos3d} extends an image detector~\cite{tian2019fcos} with additional 3D regression branches, which is later improved by \cite{wang2021probabilistic,chen2022epropnp} in depth modeling. Instead of performing object detection in the perspective view, ~\cite{wang2021detr3d,liu2022petr} design a DETR~\cite{zhu2020deformabledetr,wang2021anchordetr}-based detection head with learnable object queries in the 3D space. Inspired by the design of LiDAR-based detectors, another type of camera-only 3D perception models explicitly converts the camera features from perspective view to the bird's-eye view using a view transformer~\cite{pan2019crossview,roddick2018orthographic,roddick2020predicting,philion2020lift}. BEVDet~\cite{huang2021bevdet} and M$^2$BEV~\cite{xie2022m2bev} extends LSS~\cite{philion2020lift} and OFT~\cite{roddick2018orthographic} to 3D object detection and CaDDN~\cite{reading2021categorical} adds explicit depth estimation supervision to the view transformer. Recent research~\cite{li2022bevformer,zhou2022cross} also studies view transformation with multi-head attention.

\myparagraph{Multi-Sensor Fusion.}
Recently, multi-sensor fusion arouses significant interest among the 3D detection community. Existing approaches can be classified into \textit{proposal-level} and \textit{point-level} fusion methods. Early approach MV3D~\cite{chen2017mv3d} creates object proposals in 3D and projects the proposals to images to extract RoI features. \cite{qi2017frustum,wang2019frustum,nabati2020centerfusion} all lift image proposals into a 3D frustum. Recent work FUTR3D~\cite{chen2022futr3d} and TransFusion~\cite{bai2022transfusion} define object queries in the 3D space and fuses image features onto these proposals. All proposal-level fusion methods are \textit{object-centric} and cannot trivially generalize to other tasks such as BEV map segmentation. Point-level fusion methods, on the other hand, usually paint image semantic features onto foreground LiDAR points and perform LiDAR-based detection on the decorated point cloud inputs. As such, they are both \textit{object-centric} and \textit{geometric-centric}. Among these methods, \cite{vora2020pointpainting,wang2021pointaugmenting,yin2021multimodal,xu2021fusionpainting,chen2022autoalign} are (LiDAR) input-level decoration, while DCF~\cite{liang2018deep} and DeepFusion~\cite{li2022deepfusion} are feature-level decoration. 

% \myparagraph{Multi-Task Learning.}

% Multi-task CNNs have also been well-studied in the 2D computer vision community. ~\cite{ren2015faster,cai18cascadercnn} jointly perform object detection and instance segmentation, while \cite{he2017maskrcnn,sun2019hrnet,wang2019deep,gkioxari2018detecting} are also applicable to pose estimation and human-object interaction. Recent concurrent research M$^2$BEV~\cite{xie2022m2bev} and BEVFormer~\cite{li2022bevformer} jointly performs object detection and BEV segmentation in 3D. None of the above methods considers multi-sensor fusion. MMF~\cite{liang2019mmf} simultaneously works on depth completion and object detection with both camera and LiDAR inputs, but is still object-centric and not applicable to BEV map  segmentation. 

In contrast to all existing methods, \model performs sensor fusion in a shared BEV space and treats foreground and background, geometric and semantic information equally. It is a generic multi-task multi-sensor perception framework.
\section{Method}
\label{sect:method}

\model, as shown in \fig{fig:method:overview},  focuses on \textit{multi-sensor fusion} (\ie, multi-view cameras and LiDAR) for \textit{multi-task 3D perception} (\ie, detection and segmentation). Given different sensory inputs, we first apply modality-specific encoders to extract their features. We transform multi-modal features into a unified BEV representation that preserves both geometric and semantic information. We identify the efficiency bottleneck of the view transformation and accelerate BEV pooling with precomputation and interval reduction. We then apply the convolution-based BEV encoder to the unified BEV features to alleviate the local misalignment between different features. Finally, we append a few task-specific heads to support different 3D tasks.

% \SH{View Discrepancy is the key challenge of sensor fusion and should be highlighted in the intro. also highlight how we solve this issue. this is not answered in the paper.}

\subsection{Unified Representation}

\begin{figure*}[t]
\centering
\includegraphics[width=0.9\linewidth]{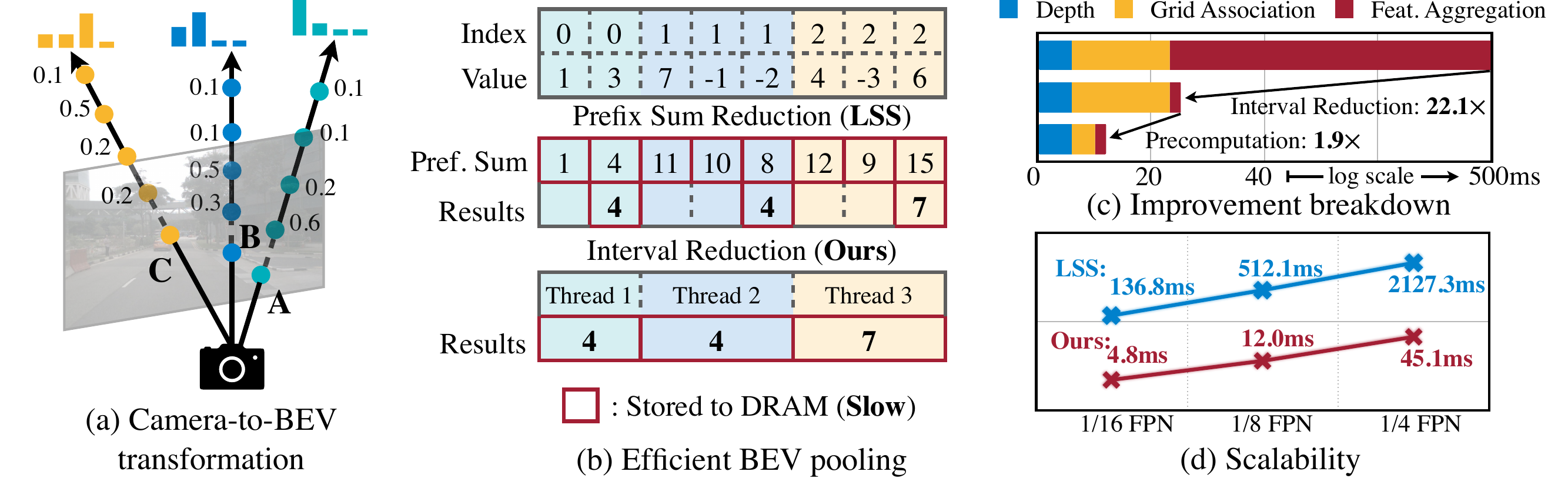}
\caption{Camera-to-BEV transformation (a) is the key step to perform sensor fusion in the unified BEV space. Existing implementation is extremely slow and takes up to 2s for a single scene. We propose efficient BEV pooling (b) using interval reduction and fast grid association with precomputation, bringing about \textbf{40$\times$} speedup to view transformation (c, d).}
\label{fig:method:vtransform}
\vspace{-15pt}
\end{figure*}

Different features can exist in different views. For instance, camera features are in the perspective view, while LiDAR/radar features are typically in the 3D/bird's-eye view. Even for camera features, each one of them has a distinct viewing angle (\ie, front, back, left, right). This \emph{view discrepancy} makes the feature fusion difficult since the same element in different feature tensors might correspond to very different spatial locations (and the \naive elementwise feature fusion will not work in this case). Thus, it is crucial to find a \emph{shared} representation, such that (1) all sensor features can be easily converted to it without information loss, and (2) it is suitable for different types of tasks.

\myparagraph{To Camera.}
Motivated by RGB-D data, one choice is to project the LiDAR point cloud to the camera plane and render the 2.5D sparse depth. However, this conversion is \emph{geometrically lossy}. Two neighbors on the depth map can be far away from each other in the 3D space. This makes the camera view less effective for tasks that focus on the object/scene geometry, such as 3D object detection.
% \SH{detection?}.

\myparagraph{To LiDAR.}
Most state-of-the-art sensor fusion methods~\cite{vora2020pointpainting,yin2021multimodal,li2022deepfusion} decorate LiDAR points with their corresponding camera features (\eg, semantic labels, CNN features or virtual points). However, this camera-to-LiDAR projection is \emph{semantically lossy}. Camera and LiDAR features have drastically different densities, resulting in only less than 5\% of camera features being matched to a LiDAR point (for a 32-channel LiDAR scanner). Giving up the semantic density of camera features severely hurts the model's performance on semantic-oriented tasks (such as BEV map segmentation). Similar drawbacks also apply to more recent fusion methods in the latent space (\eg, object query)~\cite{chen2022futr3d,bai2022transfusion}.

\myparagraph{To Bird's-Eye View.}
We \emph{adopt the bird's-eye view (BEV) as the unified representation for fusion}. This view is friendly to almost all perception tasks since the output space is also in BEV. More importantly, the transformation to BEV keeps both geometric structure (from LiDAR features) and semantic density (from camera features). On the one hand, the LiDAR-to-BEV projection flattens the sparse LiDAR features along the height dimension, thus does not create geometric distortion in \fig{fig:teaser}\textcolor{linkcolor}{a}. On the other hand, camera-to-BEV projection casts each camera feature pixel back into a ray in the 3D space (detailed in the next section), which can result in a dense BEV feature map in \fig{fig:teaser}\textcolor{linkcolor}{c} that retains full semantic information from the cameras.

\begin{table*}[t]
\setlength{\tabcolsep}{8pt}
\small\centering
\caption{BEVFusion achieves state-of-the-art 3D object detection performance on nuScenes (\texttt{val} and \texttt{test}) without bells and whistles. It breaks the convention of decorating camera features onto the LiDAR point cloud and delivers at least 1.3\% higher mAP and NDS with \textbf{1.5-2$\times$} lower computation cost. ($^{*}$: our re-implementation; $^{\dag}$: with test-time augmentation)}
\label{tab:results:detection}
\begin{tabular}{lcccccccc}
    \toprule
     & Modality & mAP (\textit{test}) & NDS (\textit{test}) & mAP (\textit{val}) & NDS (\textit{val}) & MACs (G) & Latency (ms)\\
    \midrule
    M$^2$BEV~\cite{xie2022m2bev} & C & 42.9 & 47.4 & 41.7 & 47.0 & -- & --\\
    BEVFormer~\cite{li2022bevformer} & C & 44.5 & 53.5 & 41.6 & 51.7 & -- & -- \\
    \midrule
    PointPillars~\cite{lang2019pointpillars} & L & -- & -- & 52.3 & 61.3 & 65.5 & 34.4\\
    SECOND~\cite{yan2018second} & L & 52.8 & 63.3 & 52.6 & 63.0 & 85.0 & 69.8\\
    CenterPoint~\cite{yin2021center} & L & 60.3 & 67.3 & 59.6 & 66.8 & 153.5 & 80.7\\
    %TransFusion-L~\cite{bai2022transfusion} & L & \\
    \midrule
    PointPainting~\cite{vora2020pointpainting} & C+L & -- & -- & {\kern 0.45em}65.8$^{*}$ & {\kern 0.45em}69.6$^{*}$ & 370.0 & 185.8\\
    PointAugmenting~\cite{wang2021pointaugmenting} & C+L & {\kern 0.45em}66.8$^{\dag}$ & {\kern 0.45em}71.0$^{\dag}$ & -- & -- & 408.5 & 234.4\\
    MVP~\cite{yin2021multimodal} & C+L & 66.4 & 70.5 & {\kern 0.45em}66.1$^{*}$ & {\kern 0.45em}70.0$^{*}$ & 371.7 & 187.1\\
    FusionPainting~\cite{xu2021fusionpainting} & C+L & 68.1 & 71.6 & 66.5 & 70.7 & -- & -- \\
    AutoAlign~\cite{chen2022autoalign} & C+L & -- & -- & 66.6 & 71.1 & -- & -- \\
    FUTR3D~\cite{chen2022futr3d} & C+L & -- & -- & 64.5 & 68.3 & 1069.0 & 321.4 \\
    TransFusion~\cite{bai2022transfusion} & C+L & 68.9 & 71.6 & 67.5 & 71.3 & 485.8 & 156.6\\
    \textbf{\model} (Ours) & C+L & \textbf{70.2} & \textbf{72.9} & \textbf{68.5} & \textbf{71.4} & \textbf{253.2} & \textbf{119.2}\\
    \bottomrule
\end{tabular}
\vspace{-10pt}
\end{table*}

% \haotian{We can potentially admit that using BEV as the fusion space is widely accepted in the camera-only detection community. However, the camera-only detection community does not consider efficiency, they use extremely slow view transformers (we can cite the latency of BEVFormer's view transformer and measured results for LSS). In the multi-modality case, however, since we need to also extract features from LiDAR, and the point cloud networks are slow on GPUs, we cannot afford a slow view transformer. Given the fact that the LiDAR-centric fusion does not have a slow view transformer and already achieves very good performance, this can partially explain why the community is NOT using BEV as the fusion space for multi-sensor fusion. In this paper we break this conventional understanding.}

% \haotian{ours: 2ms on 3090, LSS: 500ms on 3090 -> even slower than RTX; original: geom 11.2ms, sort 5.9ms, pool 500ms, compute 6.0ms; now geom 0ms, sort 3.0ms, pool 2.0ms, compute 6.0ms}

\subsection{Efficient Camera-to-BEV Transformation}
\label{sect:method:bev_pooling}

Camera-to-BEV transformation is non-trivial because the depth associated with each camera feature pixel is inherently ambiguous. Following LSS~\cite{philion2020lift}, we explicitly predict the discrete depth distribution of each pixel. We then scatter each feature pixel into $D$ discrete points along the camera ray and rescale the associated features by their corresponding depth probabilities (\fig{fig:method:vtransform}\textcolor{linkcolor}{a}). This generates a camera feature point cloud of size $NHWD$, where $N$ is the number of cameras and $(H, W)$ is the camera feature map size. Such 3D feature point cloud is quantized along the $x, y$ axes with a step size of $r$ (\eg, 0.4m). We use the \textit{BEV pooling} operation to aggregate all features within each $r\times r$ BEV grid and flatten the features along the $z$-axis.

Though simple, BEV pooling is surprisingly inefficient and slow, taking more than 500ms on an RTX 3090 GPU (while the rest of our model only takes around 100ms). This is because the camera feature point cloud is very large: for a typical workload\footnote{$N = 6$, $(H, W) = (32, 88)$, and $D = (60-1) / 0.5 = 118$. This corresponds to six multi-view cameras, each associated with a 32$\times$88 camera feature map (which is downsampled from a 256$\times$704 image by 8$\times$). The depth is discretized into $[1, 60]$ meters with a step size of 0.5 meter.}, there could be around 2 million points generated for each frame, two orders of magnitudes denser than a LiDAR feature point cloud. To lift this efficiency bottleneck, we propose to optimize the BEV pooling with precomputation and interval reduction.

\myparagraph{Precomputation.}
The first step of BEV pooling is to \textit{associate} each point in the camera feature point cloud with a BEV grid. Different from LiDAR point clouds, the coordinates of the camera feature point cloud are \emph{fixed} (as long as the camera intrinsics and extrinsics stay the same, which is usually the case after proper calibration). Motivated by this, we precompute the 3D coordinate and the BEV grid index of each point. We also sort all points according to grid indices and record the rank of each point. During inference, we only need to reorder all feature points based on the precomputed ranks. This caching mechanism can reduce the latency of grid association from 17ms to 4ms.

\myparagraph{Interval Reduction.}
After grid association, all points within the same BEV grid will be consecutive in the tensor representation. The next step of BEV pooling is to \textit{aggregate} the features within each BEV grid by some symmetric function (\eg, mean, max, and sum). As in \fig{fig:method:vtransform}\textcolor{linkcolor}{b}, existing implementation~\cite{philion2020lift} first computes the prefix sum over all points and then subtracts the values at the boundaries where indices change. However, the prefix sum operation requires tree reduction on the GPU and produces many unused partial sums (since we only need those values on the boundaries), both of which are inefficient. To accelerate feature aggregation, we implement a specialized GPU kernel that parallelizes directly over BEV grids: we assign a GPU thread to each grid that calculates its interval sum and writes the result back. This kernel removes the dependency between outputs (thus does not require multi-level tree reduction) and avoids writing the partial sums to the DRAM, reducing the latency of feature aggregation from 500ms to 2ms (\fig{fig:method:vtransform}\textcolor{linkcolor}{c}).

\myparagraph{Takeaways.}
The camera-to-BEV transformation is \textbf{40$\times$} faster with our optimized BEV pooling: the latency is reduced from more than 500ms to 12ms (only 10\% of our model's end-to-end runtime) and scales well across different feature resolutions (\fig{fig:method:vtransform}\textcolor{linkcolor}{d}). This is a key enabler for unifying multi-modal sensory features in the shared BEV representation. Two concurrent works of ours also identify this efficiency bottleneck in the camera-only 3D detection. They approximate the view transformer by assuming uniform depth distribution~\cite{xie2022m2bev} or truncating the points within each BEV grid~\cite{huang2021bevdet}. In contrast, our techniques are \emph{exact} without any approximation, while still being faster.

\subsection{Fully-Convolutional Fusion}
With all sensory features converted to the shared BEV representation, we can easily fuse them together with an elementwise operator (such as concatenation). Though in the same space, LiDAR BEV features and camera BEV features can still be spatially misaligned to some extent due to the inaccurate depth in the view transformer. To this end, we apply a convolution-based BEV encoder (with a few residual blocks) to compensate for such local misalignments.
% \textcolor{revisioncolor}{We also experiment with more advanced fusion mechanisms, such as gated~\cite{hu2018squeeze}, non-local~\cite{cao2019gcnet} and deformable~\cite{zhu2019deformable} fusion. However, we find that all these designs bring about at most 0.1\% improvement for the final accuracy. We believe this is because the BEV encoder (with more than 10 layers) already has enough capacity.}
Our method could potentially benefit from more accurate depth estimation (\eg, supervising the view transformer with ground-truth depth~\cite{reading2021categorical,park2021dd3d}), which we leave for future work.

% \SH{the method looks interesting. need a figure to describe. is this the solution to View Discrepancy?}
\subsection{Multi-Task Heads}
\label{sect:method:heads}

We apply multiple task-specific heads to the fused BEV feature map. Our method is applicable to most 3D perception tasks. For 3D object detection, we follow \cite{yin2021center,bai2022transfusion} to use a class-specific center heatmap head to predict the center location of all objects and a few regression heads to estimate the object size, rotation, and velocity. For 
map segmentation, different map categories may overlap (\eg, crosswalk is a subset of drivable space). Therefore, we formulate this problem as multiple binary semantic segmentation, one for each class. We follow CVT~\cite{zhou2022cross} to train the segmentation head with the standard focal loss~\cite{lin2017focal}.
\section{Experiments}
\label{sect:exp}
\begin{table}
\caption{\textcolor{revisioncolor}{BEVFusion achieves state-of-the-art 3D object detection performance among all submissions on Waymo open dataset (\texttt{test}). ($^\dag$: with test-time augmentation, $^\ddag$: with both test-time augmentation and model ensemble)}}
\label{tab:results:detection:waymo:test:ensemble}
\centering\small
\scalebox{0.9}{
\setlength{\tabcolsep}{2.0pt}
\begin{tabular}{lccccc} \toprule[1pt]
                & Frames     & mAP/L1        & mAPH/L1       & mAP/L2        & \underline{mAPH/L2} \\\midrule[0.5pt]
AFDetV2-Ens~\cite{ge2021afdet}$\ddag$ & \textbf{3} & 84.1 & 82.6 & 79.0 & 77.6 \\
InceptionLiDAR      & 10          & 83.8          & 82.5          & 79.2          & 77.8            \\
3DAL-Ens~\cite{qi2021offboard}       & 5          & 84.6          & 83.1          & 79.7          & 78.2            \\
DeepFusion-Ens~\cite{li2022deepfusion}$\ddag$  & 5          & 84.4          & 83.2          & 79.5          & 78.4             \\
MT-Net$\ddag$~\cite{chen2022mtnet}           & 3          & 84.7          & 83.2          & 79.9          & 78.5             \\
MT3D         & 4          & 85.0          & 83.7          & 80.1          & 78.7             \\
LIVOX-Detection & 7          & 84.8          & 83.5          & 80.2          & 79.0             \\
MPPNet-Ens$\ddag$~\cite{chen2022mppnet}      & 16         & 85.0          & 83.7          & 80.5          & 79.1             \\
3DAM-Ens       & 5          & 85.3          & 83.8          & 80.7          & 79.2             \\
\textbf{BEVFusion} (Ours)$\dag$  & \textbf{3} & \textbf{85.7} & \textbf{84.4} & \textbf{80.8} & \textbf{79.5}    \\

\bottomrule[1pt]
\end{tabular}
}
\vspace{-20pt}
\end{table}
\begin{table*}[t]
\setlength{\tabcolsep}{9pt}
\small\centering
\caption{BEVFusion outperforms the state-of-the-art multi-sensor fusion methods by \textbf{13.6\%} on BEV map segmentation on nuScenes (\texttt{val}) with consistent improvements across different categories.}
\label{tab:results:segmentation}
\begin{tabular}{lcccccccc}
    \toprule
     & Modality & Drivable & Ped. Cross. & Walkway & Stop Line & Carpark & Divider & Mean \\
    \midrule
    OFT~\cite{roddick2018orthographic} & C & 74.0 & 35.3 & 45.9 & 27.5 & 35.9 & 33.9 & 42.1 \\
    LSS~\cite{philion2020lift} & C & 75.4 & 38.8 & 46.3 & 30.3 & 39.1 & 36.5 & 44.4 \\
    CVT~\cite{zhou2022cross} & C & 74.3 & 36.8 & 39.9 & 25.8 & 35.0 & 29.4 & 40.2 \\
    M\textsuperscript{2}BEV~\cite{xie2022m2bev} & C & 77.2 & -- & -- & -- & -- & 40.5 & --\\
    \textbf{BEVFusion} (Ours) & C & \textbf{81.7} & \textbf{54.8} & \textbf{58.4} & \textbf{47.4} & \textbf{50.7} & \textbf{46.4} & \textbf{56.6} \\
    \midrule
    PointPillars~\cite{lang2019pointpillars} & L & 72.0 & 43.1 & 53.1 & 29.7 & 27.7 & 37.5 & 43.8 \\
    CenterPoint~\cite{yin2021center} & L & 75.6 & 48.4 & 57.5 & 36.5 & 31.7 & 41.9 & 48.6 \\
    \midrule
    PointPainting~\cite{vora2020pointpainting} & C+L & 75.9 & 48.5 & 57.1 & 36.9 & 34.5 & 41.9 & 49.1\\
    MVP~\cite{yin2021multimodal} & C+L & 76.1 & 48.7 & 57.0 & 36.9 & 33.0 & 42.2 & 49.0\\
    \textbf{BEVFusion} (Ours) & C+L & \textbf{85.5} & \textbf{60.5} & \textbf{67.6} & \textbf{52.0} & \textbf{57.0} & \textbf{53.7} & \textbf{62.7} \\
    \bottomrule
\end{tabular}
\vspace{-5pt}
\end{table*}
\begin{table*}[h]
\centering\small
\setlength{\tabcolsep}{8pt}
\caption{BEVFusion is robust under different lighting and weather conditions, significantly boosting the performance single-modality models under challenging rainy\textsubscript{\textcolor{mydarkgreen}{(+10.7)}} and nighttime\textsubscript{\textcolor{mydarkgreen}{(+12.8)}} scenes.}
\label{tab:results:analysis:weather_and_lighting}
\scalebox{0.93}{
\begin{tabular}{lccccccccc}\toprule
& & \multicolumn{2}{c}{Sunny} & \multicolumn{2}{c}{Rainy} & \multicolumn{2}{c}{Day} & \multicolumn{2}{c}{Night} \\
\cmidrule(r){3-4}\cmidrule(r){5-6}\cmidrule(r){7-8}\cmidrule(r){9-10}
& Modality & mAP & mIoU & mAP & mIoU & mAP & mIoU & mAP & mIoU \\\midrule
CenterPoint~\cite{yin2021center} & L  & 62.9 & 50.7 & 59.2 & 42.3 & 62.8 & 48.9 & 35.4 & 37.0 \\
%\textcolor{revisioncolor}{TransFusion-L}~\cite{bai2022transfusion} & \textcolor{revisioncolor}{L} & \textcolor{revisioncolor}{64.5} & \textcolor{revisioncolor}{--} & \textcolor{revisioncolor}{64.3} & \textcolor{revisioncolor}{--} & \textcolor{revisioncolor}{69.5} & \textcolor{revisioncolor}{--} & \textcolor{revisioncolor}{36.5} & \textcolor{revisioncolor}{--}\\
\textcolor{revisioncolor}{BEVFormer}~\cite{li2022bevformer} & \textcolor{revisioncolor}{C} & \textcolor{revisioncolor}{41.0} & \textcolor{revisioncolor}{--} & \textcolor{revisioncolor}{44.0} & \textcolor{revisioncolor}{--} & \textcolor{revisioncolor}{41.9} & \textcolor{revisioncolor}{--} & \textcolor{revisioncolor}{21.2} & \textcolor{revisioncolor}{--}\\
BEVFusion & C & -- & 59.0 & -- & 50.5 & -- & 57.4 & -- & 30.8 \\
\midrule
MVP & C+L   & 65.9 \textsubscript{(+3.0)} & 51.0 \textcolor{mydarkgray}{\textsubscript{(+0.3)}} & 66.3 \textcolor{mydarkblue}{\textsubscript{(+7.1)}} & 42.9 \textcolor{mydarkgray}{\textsubscript{(+0.6)}} & 66.3 \textsubscript{(+3.5)} & 49.2 \textcolor{mydarkgray}{\textsubscript{(+0.3)}} & 38.4 \textsubscript{(+3.0)} & 37.5 \textcolor{mydarkgray}{\textsubscript{(+0.5)}} \\
BEVFusion & C+L  & 68.2 \textcolor{mydarkblue}{\textsubscript{(+5.3)}} & 65.6 \textcolor{mydarkblue}{\textsubscript{(+6.6)}} & 69.9 \textcolor{mydarkgreen}{\textsubscript{(+10.7)}} & 55.9 \textcolor{mydarkblue}{\textsubscript{(+5.4)}} & 68.5 \textcolor{mydarkblue}{\textsubscript{(+5.7)}} & 63.1 \textcolor{mydarkblue}{\textsubscript{(+5.7)}} & 42.8 \textcolor{mydarkblue}{\textsubscript{(+7.4)}} & 43.6 \textcolor{mydarkgreen}{\textsubscript{(+12.8)}} \\
\bottomrule
\end{tabular}
}
\vspace{-15pt}
\end{table*}

We evaluate BEVFusion for camera-LiDAR fusion on 3D object detection and BEV map segmentation, covering both geometric- and semantic-oriented tasks. Our framework can be easily extended to support other types of sensors (such as radars and event-based cameras) and other 3D perception tasks (such as 3D object tracking and motion forecasting).

\myparagraph{Model.}
We use Swin-T~\cite{liu2021Swintransformer} as our image backbone and VoxelNet~\cite{yan2018second} as our LiDAR backbone. We apply FPN~\cite{lin2017feature} to fuse multi-scale camera features to produce a feature map of 1/8 input size. We downsample camera images to 256$\times$704 and voxelize the LiDAR point cloud with 0.075m (for detection) and 0.1m (for segmentation). As detection and segmentation tasks require BEV feature maps with different spatial ranges and sizes, we apply grid sampling with bilinear interpolation before each task-specific head to explicitly transform between different BEV feature maps.

%\myparagraph{Training.}
%We pre-train the image backbone on nuImage (for 2D object detection) following~\cite{vora2020pointpainting,wang2021pointaugmenting,yin2021multimodal,bai2022transfusion}. Unlike existing approaches~\cite{vora2020pointpainting,wang2021pointaugmenting,bai2022transfusion} that freeze the camera encoder, we train the entire model in an end-to-end manner. We apply both image and LiDAR data augmentations to prevent overfitting.  Optimization is carried out using AdamW~\cite{loshchilov2017decoupled} with a weight decay of $10^{-2}$. We refer the readers to the appendix for more details \textcolor{revisioncolor}{on training schedules and data augmentation}.

\myparagraph{Dataset.}
We evaluate our method on nuScenes~\cite{caesar2020nuscenes} and Waymo~\cite{sun2020scalability}, which are large-scale datasets for 3D perception with $>$40k annotated scenes. Each sample in both datasets are equipped with both LiDAR and surrounding camera inputs.  

\subsection{3D Object Detection}

%\zhijian{Need a paragraph here}

We first experiment on the geometric-centric 3D object detection benchmark, where BEVFusion achieves superior performance with lower computation cost and measured latency. We use the mean average precision (mAP) across 10 foreground classes and the nuScenes detection score (NDS) as our detection metrics. We also measure the single-inference \#MACs and latency on an RTX3090 GPU for all open-source methods. We use a single model without any test-time augmentation for both \texttt{val} and \texttt{test} results.

% \myparagraph{Results.}

As in \tab{tab:results:detection}, BEVFusion achieves state-of-the-art results on the nuScenes detection benchmark, with close-to-real-time (\textbf{8.4 FPS}) inference speed on a desktop GPU. Compared with TransFusion~\cite{bai2022transfusion}, BEVFusion offers 1.3\% improvement in \texttt{test} split mAP and NDS, while significantly reduces the MACs by \textbf{1.9$\times$} and measured latency by \textbf{1.3$\times$}. It also compares favorably against representative point-level fusion methods PointPainting~\cite{vora2020pointpainting} and MVP~\cite{yin2021multimodal} with \textbf{1.6$\times$} speedup, \textbf{1.5$\times$} MACs reduction and \textbf{3.8\%} higher mAP on the \texttt{test} set. We argue that the efficiency gain of BEVFusion comes from the fact that we choose the BEV space as the shared fusion space, which fully utilizes all camera features instead of just a 5\% sparse set. Consequently, BEVFusion can achieve the same performance with much smaller resolution for the camera inputs, resulting in significantly lower MACs. Combined with the efficient BEV pooling operator in \sect{sect:method:bev_pooling}, BEVFusion transfers MACs reduction into measured speedup. 

BEVFusion also achieves state-of-the-art performance on the Waymo open dataset~\cite{sun2020scalability} (\tab{tab:results:detection:waymo:test:ensemble}). BEVFusion outperforms the previous state-of-the-art multi-modal detector, DeepFusion~\cite{li2022deepfusion} with 60\% of input frames. Furthermore, DeepFusion ensembles 25 models evaluated with test-time augmentation, while we deliver better performance by applying test-time augmentation to a single BEVFusion model. 

\begin{figure*}[t]
\centering
\subfloat[Different object sizes]{\includegraphics[width=0.3\linewidth]{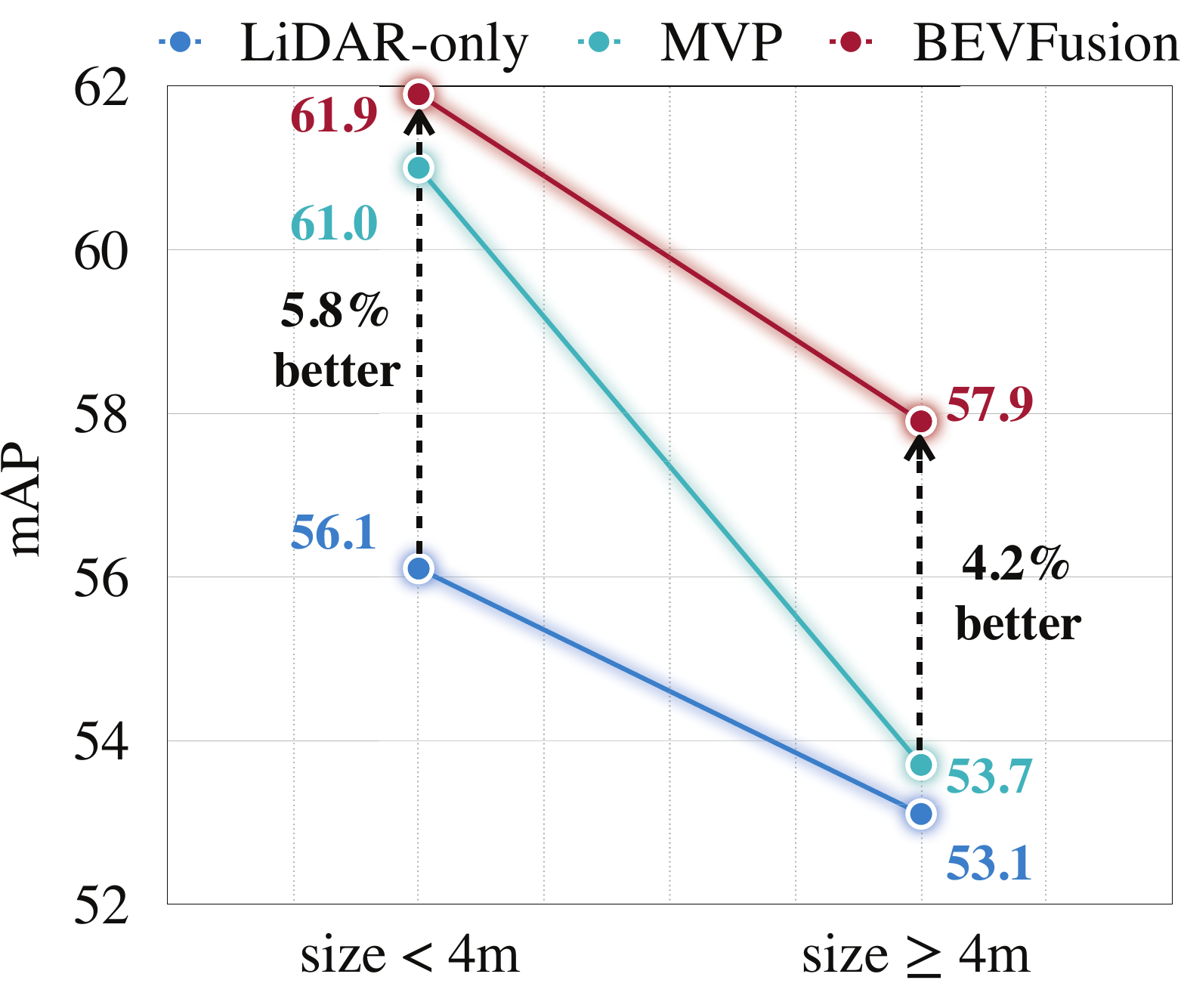}\label{fig:analysis:sizes}}
\hfill
\subfloat[Different object distances]{\includegraphics[width=0.3\linewidth]{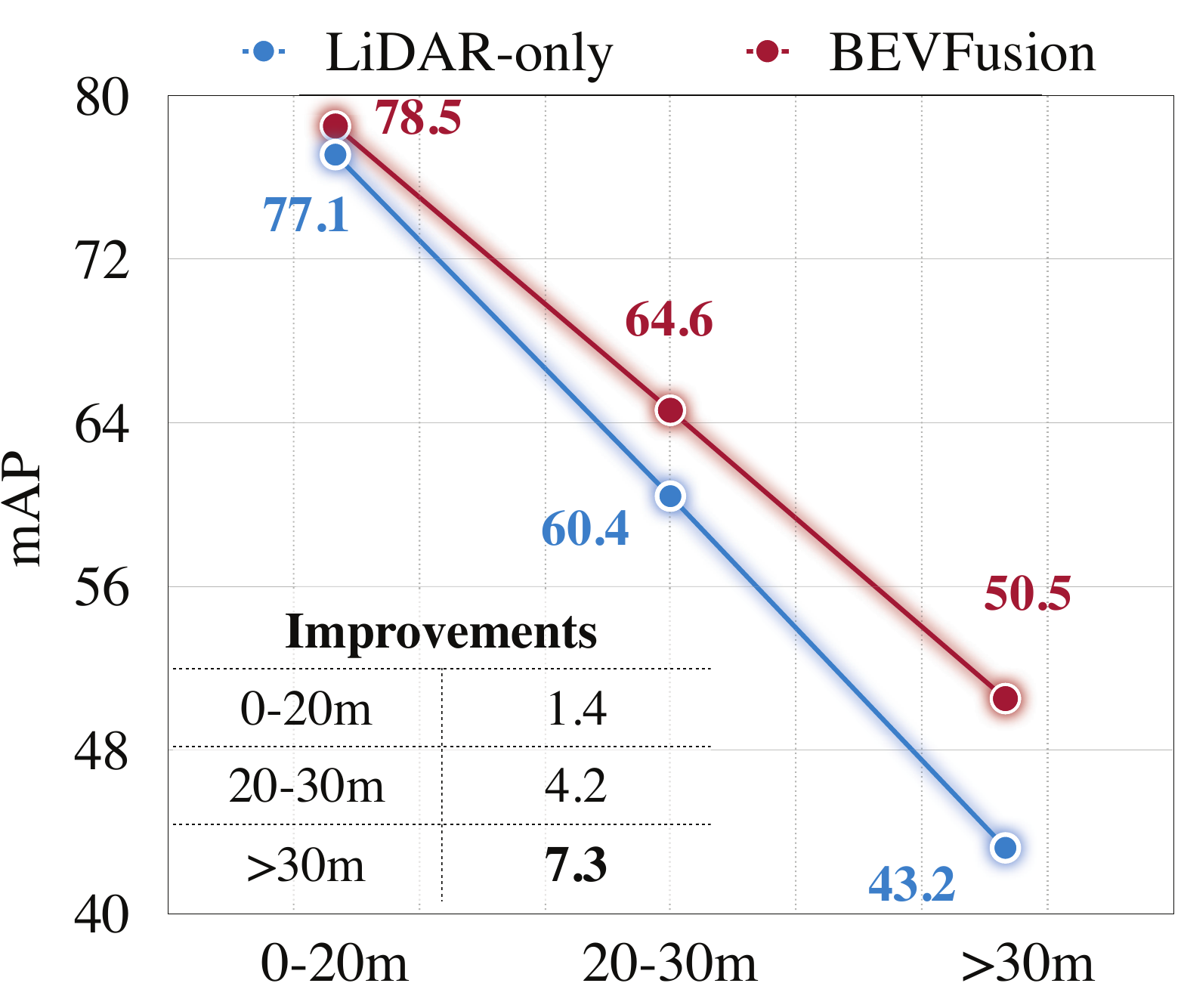}\label{fig:analysis:distances}}
\hfill
\subfloat[Different LiDAR sparsity]{\includegraphics[width=0.3\linewidth]{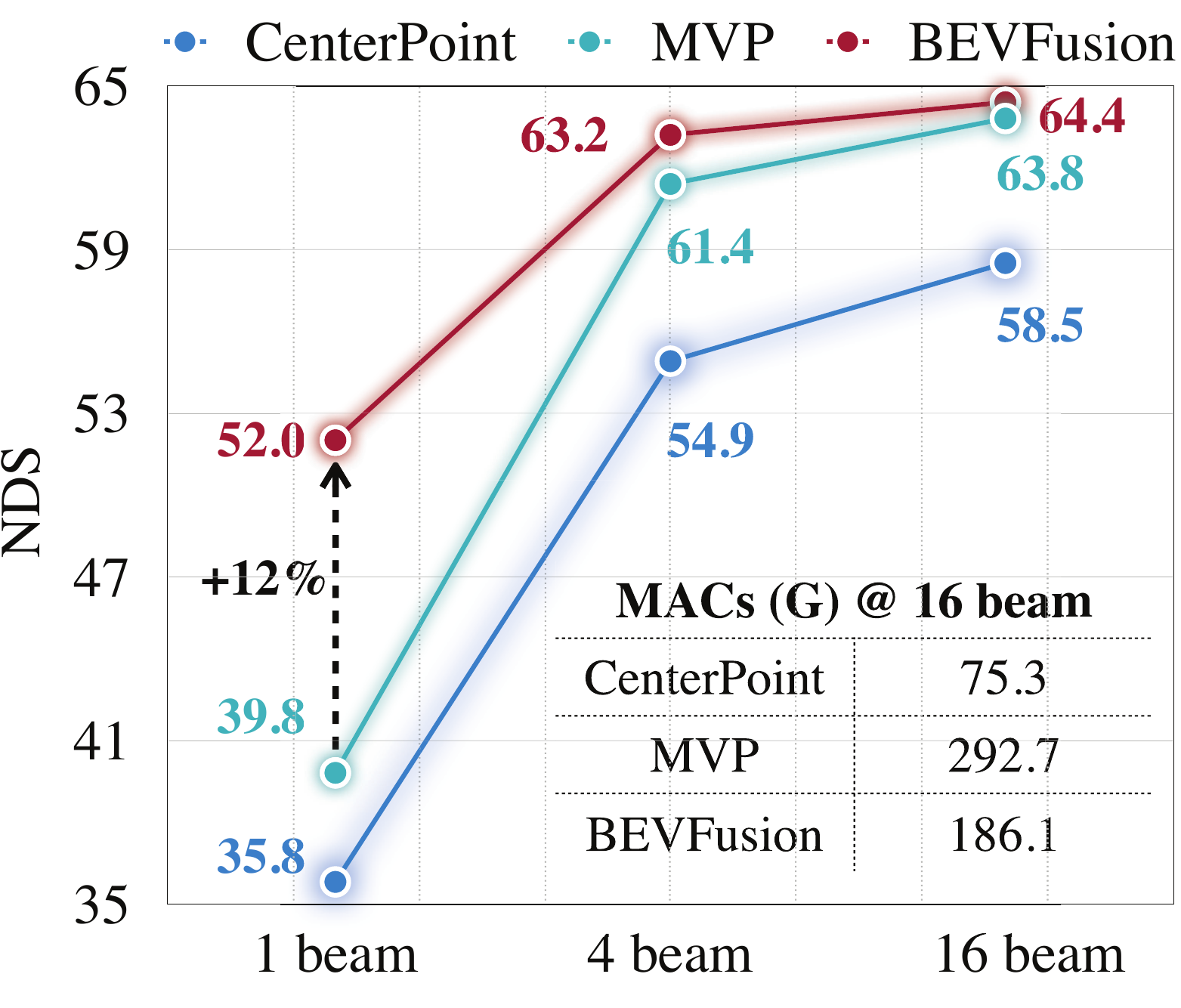}\label{fig:analysis:sparse}}
\caption{BEVFusion outperforms state-of-the-art single- and multi-modality detectors under different LiDAR sparsity, object sizes and object distances, especially under more challenging settings (\ie, \textit{sparser point clouds}, \textit{small/distant objects}).}
\label{fig:analysis}
\vspace{-20pt}
\end{figure*}

\subsection{BEV Map Segmentation}

We further compare BEVFusion with state-of-the-art models on the semantic-centric BEV map segmentation task, where BEVFusion achieves an even larger performance boost. We report the Intersection-over-Union (IoU) on 6 background classes and the class-averaged mean IoU as our evaluation metric. As different classes may have overlappings (\eg car-parking area is also drivable), we evaluate the {binary segmentation} performance for each class separately and select the highest IoU across different thresholds~\cite{zhou2022cross}. For each frame, we only perform the evaluation in the [-50m, 50m]$\times$[-50m, 50m] region around the ego car following~\cite{philion2020lift,zhou2022cross,xie2022m2bev,li2022bevformer}.

% In BEVFusion, we use a single model that jointly performs binary segmentation for all classes instead of following the conventional approach to train a separate model for each class. This results in \textbf{6$\times$} faster inference and training. We reproduced the results of all open-source competing methods.

% \myparagraph{Results.}

We report the BEV map segmentation results in \tab{tab:results:segmentation}. In contrast to 3D object detection which is a \textit{geometric}-oriented task, map segmentation is \textit{semantic}-oriented. As a result, our camera-only BEVFusion model outperforms LiDAR-only baselines by \textbf{8-13\%}. This observation is the exact opposite of results in \tab{tab:results:detection}, where state-of-the-art camera-only 3D detectors got outperformed by LiDAR-only detectors by almost 20 mAP. Our camera-only model boosts the performance of existing monocular BEV map segmentation methods by at least \textbf{12\%}. In the multi-modality setting, we further improve the performance of the monocular BEVFusion by \textbf{6} mIoU and achieved $>$\textbf{13\%} improvement over state-of-the-art sensor fusion methods~\cite{vora2020pointpainting,yin2021multimodal}. This is because both baseline methods are \textit{object}-centric and \textit{geometric}-oriented. PointPainting~\cite{vora2020pointpainting} only decorates the \textit{foreground} LiDAR points and MVP only densifies \textit{foreground} 3D objects. Both approaches are not helpful for segmenting map components. Worse still, both methods assume that LiDAR should be the more effective modality in sensor fusion, which is not true according to our observations in \tab{tab:results:segmentation}.

\section{Analysis}

We present in-depth analyses of BEVFusion over single-modality models and state-of-the-art multi-modality models.

\myparagraph{Weather and Lighting.}
We first analyze the performance of BEVFusion under different weather and lighting conditions in \tab{tab:results:analysis:weather_and_lighting}.
LiDAR-only models face significant challenges in detecting objects in rainy weather due to sensor noise, while BEVFusion leverages the robustness of camera sensors to achieve a \textbf{10.7} mAP improvement, which largely narrows the performance gap between sunny and rainy scenarios. Poor lighting conditions pose challenges for both detection and segmentation models. For detection, MVP's improvement is relatively small compared to BEVFusion, which relies less on \textit{accurate} 2D instance segmentations to generate virtual points and therefore performs better in dark or overexposed scenes. For segmentation, while camera-only BEVFusion outperforms CenterPoint on the entire benchmark, its performance drops significantly at nighttime. However, multi-modal BEVFusion achieves a \textbf{12.8} mIoU improvement, even greater than its improvement in the daytime, demonstrating the importance of leveraging geometric clues when camera sensors fail.
% Detecting objects in rainy weather is challenging for LiDAR-only models due to significant sensor noises. Thanks to the robustness of camera sensors under different weathers, BEVFusion improves CenterPoint by \textbf{10.7} mAP, closing the performance gap between sunny and rainy scenarios. Poor lighting conditions are challenging for both detection and segmentation models. For detection, MVP achieves a much smaller improvement compared to BEVFusion since it requires \textit{accurate} 2D instance segmentations to generate virtual point generation. This can be very challenging in dark or overexposed scenes. For segmentation, even if the camera-only BEVFusion greatly outperforms CenterPoint on the entire dataset in \tab{tab:results:segmentation}, its performance is much worse at nighttime. Our BEVFusion significantly boosts its performance by \textbf{12.8} mIoU,  which is even larger than the improvement in the daytime, demonstrating the significance of geometric clues when camera sensors fail.

\myparagraph{Sizes and Distances.}
We also analyze the performance of BEVFusion under different object sizes and distances. From \fig{fig:analysis:sizes}, BEVFusion achieves consistent improvements over its LiDAR-only counterpart for both small and large objects, while MVP has only negligible improvements for objects larger than 4m. This is because larger objects are typically much denser, benefiting less from augmented multi-modal virtual points (MVPs). Additionally, BEVFusion yields greater improvements to the LiDAR-only detector for smaller objects (in \fig{fig:analysis:sizes}) and more distant objects (in \fig{fig:analysis:distances}), both of which are inadequately captured by LiDAR and can therefore derive more benefit from the dense camera information.

\myparagraph{Sparser LiDARs.}
We finally demonstrate the performance of CenterPoint~\cite{yin2021center} (LiDAR-only), MVP~\cite{yin2021multimodal} (multi-modal), and our BEVFusion under different LiDAR sparsities in \fig{fig:analysis:sparse}. BEVFusion consistently outperforms MVP across all sparsity levels with a \textbf{1.6$\times$} reduction in \#MACs and achieves a \textbf{12\%} improvement in the 1-beam LiDAR scenario. MVP decorates the point cloud and directly applies CenterPoint on the painted and densified LiDAR input. As a result, it naturally requires the LiDAR-only detector (CenterPoint) to perform well, which is not valid under sparse LiDAR settings (\ie, 35.8 NDS with 1-beam input in \fig{fig:analysis:sparse}). In contrast, BEVFusion integrates multi-sensor information in the shared BEV space and does not rely solely on a robust LiDAR-only detector.

\section{Conclusion}
\label{sect:conclusion}

We present BEVFusion, an efficient and generic framework for multi-task multi-sensor 3D perception. BEVFusion unifies camera and LiDAR features in a shared BEV space that fully preserves geometric and semantic information. To achieve this, we accelerate the slow camera-to-BEV transformation by more than 40 times. BEVFusion rethinks the effectiveness of point-level fusion in multi-sensor perception systems and achieves superior performance on both nuScenes 3D detection and BEV map segmentation tasks with 1.5-1.9$\times$ less computation and 1.3-1.6$\times$ measured speedup over existing solutions. BEVFusion also outperforms all existing sensor fusion methods on Waymo open dataset. We hope that BEVFusion can serve as a simple but powerful baseline to inspire future research on multi-task multi-sensor fusion.

%\myparagraph{Limitations.}

%At present, \model still has performance degradation in joint multi-task training, which has not yet unlocked the potential for larger inference speedup in the multi-task setting. More accurate depth estimation~\cite{reading2021categorical,park2021dd3d} is also an under-explored direction in this paper that can potentially boost the performance of \model further.

%\myparagraph{Societal Impacts.}

%Efficient and accurate multi-sensor perception is crucial for the safety of autonomous vehicles. BEVFusion reduces the computation cost of state-of-the-art multi-sensor fusion models by half and achieves large accuracy improvements on small and distant objects, and in rainy and night conditions. However, it still takes a fairly large amount of computation resources (more than 200 GPU hours) to train, which could result in a large carbon footprint. Improving the training efficiency will be an important future direction.

\bibliographystyle{IEEEtran}
\bibliography{reference}

% \appendix
% \renewcommand{\thesection}{A.\arabic{section}}
% \renewcommand{\thefigure}{A\arabic{figure}}
% \renewcommand{\thetable}{A\arabic{table}}/
% \setcounter{section}{0}
% \setcounter{figure}{0}
% \setcounter{table}{0}

% \clearpage
% \input{text/appendix}

\end{document}